\documentclass[10pt,twocolumn,letterpaper]{article}

\usepackage{iccv}
\usepackage{times}
\usepackage{epsfig}
\usepackage{graphicx}
\usepackage{amsmath}
\usepackage{amssymb}

\usepackage{url}            % simple URL typesetting
\usepackage{booktabs}       % professional-quality tables
\usepackage{amsfonts}       % blackboard math symbols
\usepackage{nicefrac}       % compact symbols for 1/2, etc.
\usepackage{microtype}      % microtypography
\usepackage{lipsum}
\usepackage{xcolor}
\usepackage{caption}
\usepackage{subcaption}

\usepackage[pagebackref=true,breaklinks=true,letterpaper=true,colorlinks,bookmarks=false]{hyperref}
\iccvfinalcopy % *** Uncomment this line for the final submission

\def\iccvPaperID{48} % *** Enter the ICCV Paper ID here

\ificcvfinal\fi

\begin{document}

%%%%%%%%% TITLE
\title{NADS-Net: A Nimble Architecture for Driver and Seat Belt Detection via Convolutional Neural Networks}

% \author{
%   Sehyun Chun\\
%   Visual Intelligence Laboratory\\
%   University of Iowa\\
%   Iowa City, IA 52242\\
%   {\tt\small sehyun-chun@uiowa.edu} \\
%   \And
%   Nima Hamidi Ghalehjegh\\
% %   Visual Intelligence Laboratory\\
% %   University of Iowa\\
% %   Iowa City, IA 52242\\
% %   \texttt{nima-hamidi@uiowa.edu}\\
%   \And
%   Joseph B. Choi \\
% %   Visual Intelligence Laboratory\\
% %   University of Iowa\\
% %   Iowa City, IA 52242\\
% %   \texttt{boogun-choi@uiowa.edu} \\
%   \And
%   Stephen S. Baek%\thanks{Use footnote for providing further
% %     information about author (webpage, alternative
% %     address)---\emph{not} for acknowledging funding agencies.} \\
% %   National Advanced Driving Simulator (NADS)\\
% %   Visual Intelligence Laboratory\\
% %   University of Iowa\\
% %   Iowa City, IA 52242 \\
% %   \texttt{stephen-baek@uiowa.edu} \\https://www.overleaf.com/project/5cf662f2af222b604740aca8
% }
\author{Sehyun Chun$^1$ \and Nima Hamidi Ghalehjegh$^1$ \and Joseph B. Choi$^1$ \and Chris W. Schwarz$^2$\\
\and John G. Gaspar$^2$ \and Daniel V. McGehee$^{1,2}$ \and Stephen S. Baek$^{1,2}$\thanks{Corresponding author: stephen-baek@uiowa.edu}\\
%{\tt\small \{sehyun-chun, nima-hamidi, boogun-choi, chris-schwarz, john-gaspar, daniel-mcgehee, stephen-baek\}@uiowa.edu}
% For a paper whose authors are all at the same institution,
% omit the following lines up until the closing ``}''.
% Additional authors and addresses can be added with ``\and'',
% just like the second author.
% To save space, use either the email address or home page, not both
% \and
% Second Author\\
% Institution2\\
% First line of institution2 address\\
% {\tt\small secondauthor@i2.org}
\and
$^1$Department of Industrial and Systems Engineering, University of Iowa, Iowa City, IA 52242\\
$^2$National Advanced Driving Simulator (NADS), Iowa City, IA 52242\\
}

\maketitle
% Remove page # from the first page of camera-ready.
\ificcvfinal\thispagestyle{empty}\fi

%%%%%%%%% ABSTRACT

\begin{abstract}
A new convolutional neural network (CNN) architecture for 2D driver/passenger pose estimation and seat belt detection is proposed in this paper. The new architecture is more nimble and thus more suitable for in-vehicle monitoring tasks compared to other generic pose estimation algorithms. The new architecture, named NADS-Net, utilizes the feature pyramid network (FPN) backbone with multiple detection heads to achieve the optimal performance for driver/passenger state detection tasks. The new architecture is validated on a new data set containing video clips of 100 drivers in 50 driving sessions that are collected for this study. The detection performance is analyzed under different demographic, appearance, and illumination conditions. The results presented in this paper may provide meaningful insights for the autonomous driving research community and automotive industry for future algorithm development and data collection.
\end{abstract}

%%%%%%%%% BODY TEXT
\section{Introduction}

A vast majority of vehicle accidents reported worldwide are caused by distracted driving behaviors \cite{crash2005}. Examples of distracted driving behaviors include use of smartphones/mobile devices, smoking tobaccos, engaging in a conversation with other passengers, drinking beverages, eating foods, and such, that are irrelevant to the task of driving itself. Different forms of distractions such as drowsiness, fatigues, medication effects, and other medical/physiological issues can also cause life threatening situations \cite{fatigues2005}. 

Another significant automotive safety hazard is caused by improper/non-use of seat belt, which can cause a serious injury and fatality. According to the U.S. National Highway Traffic Safety Administration (NHTSA), 10,428 unbuckled drivers and passengers died in 2016 on the U.S. roads \cite{national2017usdot}. Moreover, even if the drivers and passengers are buckled, improperly positioned seat belt can cause fatal injuries. According to \cite{fatalInjury1998}, fatal injuries such as intra-abdominal injury are caused by improper positioning of seat belt at the time of crash.

%However, currently there is no published work that does seat belt segmentation or both seat belt segmentation together with pose estimation. In this work, we append extra branch in parallel with posture detection to simultaneously segment seat belt and estimation human key points. ({\color{red}This paragraph needs to be placed in the introduction section.})

To this end, \textit{in-vehicle monitoring systems} (IVMS) are rapidly becoming a standard technology in consumer vehicles as they can play a critical role in preventing and mitigating traffic accidents by alerting the distracted driver and adaptively adjusting the safety mechanisms. Furthermore, in the upcoming era of autonomous driving, IVMS technologies are expected to be even more critical \cite{ivms2010}. For example, an IVMS can provide an alert to the driver when the vehicle system detects an anomaly while in an autonomous driving mode, so that the driver can take over the control prior to the system failure \cite{ivmsTakeover2017}. An IVMS could also provide personalized accommodation to the occupants to maximize the comfort and safety.

For IVMS, vision-based sensing technologies are at the core, as they permit non-invasive, non-obstructive means to monitor and detect in-cabin activities. To this end, visual cues from face, eye-gaze, head-pose, hand gestures, and body poses are detected and tracked in IVMS systems \cite{Choi2016,Mbouna2013,Yan2016,Yuen2018,Okuno2018}. The goal of vision-based sensing typically includes recognition of a variety of states, activities, and aspects of human automobile users, such as the body posture of the driver and the front row passenger, and correct donning of seat belt, which are the main objectives of this paper. More specifically, this paper proposes a new convolutional neural network (CNN) architecture for 2D driver/passenger pose estimation and seat belt detection that is more nimble and, thus, more suitable for in-vehicle monitoring tasks compared to other state-of-the-art approaches. The new architecture, named \emph{NADS-Net}, utilizes a feature pyramid network (FPN) \cite{FPN2017} backbone with multi-branch detection heads, namely, a key point detection head, a part-affinity field \cite{cao2017paf} detection head, and a seat belt segmentation head. The new architecture shows similar detection accuracy in the body pose estimation task compared to the state-of-the-art algorithm \cite{cao2017paf}, while being more concise and efficient, and capable of doing more (\ie seat belt detection). In addition, we also collected a video data set of 100 drivers in 50 driving sessions to fine-tune the performance of the proposed model pre-trained on the generic human pose estimation in the wild data sets. We analyzed the performance of the new NADS-Net algorithm as well as one of the current state of the art algorithm proposed by Cao \etal \cite{cao2017paf} under different demographic, appearance, and lighting conditions. This may provide insights for future algorithm design and data collection to the academic research community and the automotive industry. The major contribution of this paper is summarized as follows:
\begin{itemize}
    \item A new architecture for driver/passenger pose estimation and seat belt detection is proposed.
    \item Insights for CNN algorithm design are distilled by contrasting the new architecture with other typical generic pose detection algorithm.
    \item Performance of the algorithms are analyzed on different imaging conditions, providing new insights and guidelines for future algorithm development and data collection.
\end{itemize}

\section{Related works}

\subsection{Human pose estimation}

In the automotive industry, human pose estimation algorithms have gained an increasing interest for their enhanced capacity in capturing kinematic posture of people without any sensor instrumentation. The taxonomy of human pose estimation methods in literature can be broadly categorized into \textit{top-down} approaches and \textit{bottom-up} approaches.

\paragraph{Top-down approaches} Top-down approaches detect person bounding boxes first and then break each bounding box down into body key points and a skeleton. 
Papandreou \etal \cite{papandreou2017} employed Faster R-CNN \cite{Ren2015FasterRT} to first predict person bounding boxes and then utilized the residual network (ResNet) \cite{he2016deep} to predict both dense heat maps and offset vectors within each person bounding box to localize key points. He \etal \cite{maskRCNN2017} proposed Mask R-CNN which extends Faster R-CNN to support both person instance segmentation and human key point detection, on top of the Faster R-CNN's bounding box detection head. Moreover, they changed the network backbone to FPN \cite{FPN2017}, which resulted performance gain in both accuracy and speed. Chen \etal \cite{cascaded2018} proposed cascaded pyramid network (CPN) comprised of two stages: GlobalNet and RefineNet. The CPN first detects the bounding box of a person and the cropped bounding box is passed to GlobalNet where key points are predicted with an FPN backbone. RefineNet then refines the key points predicted by GlobalNet, which, in turn, achieves more precise detection of occluded or invisible key points. 

\paragraph{Bottom-up approaches} Bottom-up approaches detect all body key points individually, first, and then parse their connections and memberships to construct person instances (\ie skeletons). DeepCut \cite{deepcut2016} is an example of a bottom-up approach that detects body parts and the relations between each body parts. These outputs are then used to regress the spatial offsets of detected parts and to connect person instances. Later, DeeperCut \cite{DepperCut2016} redesigned the original DeepCut algorithm by utilizing a deeper ResNet architecture to improve the body part detectors and to induce stronger pairwise scores between the body parts. However, both DeepCut and DeeperCut could not achieve a practical inference speed for real-time applications. Newell \etal \cite{Newell2017} introduced a method that can simultaneously output key point locations and pixel-wise embeddings to automatically group key point detection results into individual poses. Cao \etal \cite{cao2017paf} proposed part affinity fields (PAF) that encompass vector fields indicating how individual key points should be connected. They augmented the convolution pose machine \cite{cpm2016} algorithm with the PAF prediction head and employed bipartite graph matching to greedily parse skeleton instances.

%{\color{red}In this work, we make use of Part Afinity Field \cite{cao2017paf}, which is a bottom-up approach that can run in real-time, for in-vehicle posture estimation task. Cao \cite{cao2017paf} utilized VGG network\cite{Simonyan2014VeryDC} as a backbone with 6 multiple refining stages. However, it is observed that both He \etal \cite{maskRCNN2017} and Chen \etal \cite{cascaded2018} have gained performance boost from using FPN. Hence, our work implement FPN backbone instead of VGG backbone to replace 6 multiple refining stages to increase speed while keeping the accuracy the same.}

\paragraph{In-vehicle human pose estimation} Despite the fact that there have been significant breakthroughs in generic pose estimation tasks, there are only few pose estimation models in literature specifically for in-vehicle use. For example, Okuno \etal \cite{Okuno2018} proposed a method that predicts both human posture and face orientation in real-time. Unlike other generic posture estimation models, their model relies on a relatively shallow CNN architecture, comprised of only three convolution layers and two succeeding fully connected layers that directly regress $x$ and $y$ coordinates of eight body parts and face orientation angle. Their model was trained and evaluated on a custom data set comprised of images of twelve subjects, collected for their study. Yuen \etal \cite{Yuen2018} presented a model which modified the PAF model of Cao \etal \cite{cao2017paf} that only focuses on detecting the wrists and elbows of both the driver and the front passenger. They also collected their own data set that consists of real on-road scenes with varying lighting conditions to train and test the model. The method proposed in this paper substantially expands these previous works towards more comprehensive and reliable detection performance.

\begin{figure*}
\centering
    \includegraphics[width=\linewidth]{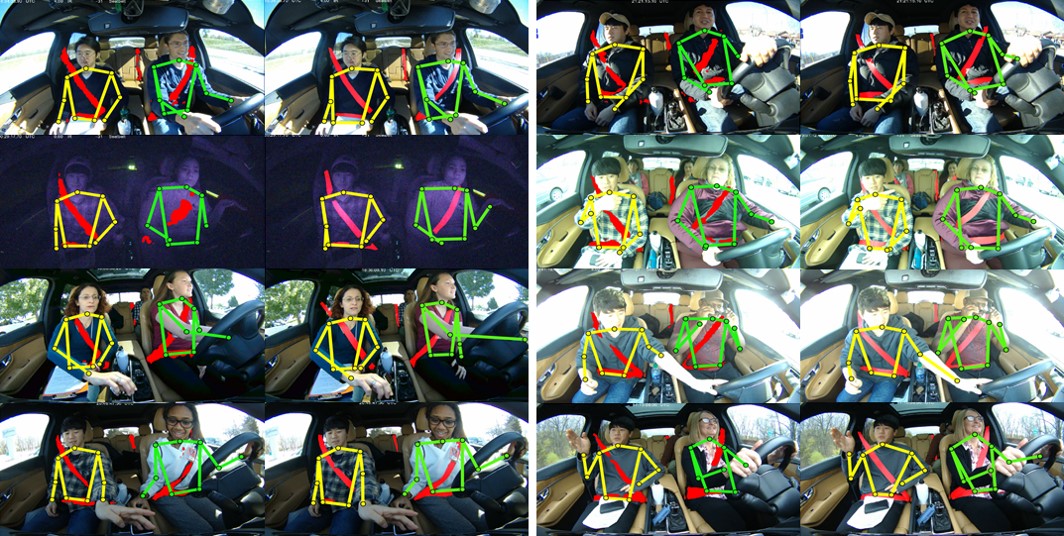}
    
    \caption{First and third columns in the figure are results produced by NADS-Net and second and fourth columns are the corresponding ground truth annotations.}
    \label{fig:ICCV_gt_img}
\end{figure*}

\subsection{Seat belt}

There have been ongoing efforts regarding computer vision-based detection of seat belt use. Zhou \etal \cite{zhou2017_1} combined an edge detection method, the salient gradient map, and the radial basis functions (RBF) into a unified network architecture to identify whether there is seat belt present in the image or not. Guo \etal \cite{Guo2011} similarly utilized an edge detection algorithm to detect seat belt from traffic surveillance cameras. Zhou \etal \cite{zhou2017_2} used AlexNet \cite{Krizhevsky2012ImageNetCW} with batch normalization \cite{Ioffe2015BatchNA} to identify seat belts. Elihos \etal \cite{elihos2018} proposed a method that crops passenger regions first using the single shot detector (SSD) \cite{Liu2016SSDSS} and applies a CNN to detect seat belt non-use. The seat belt detection algorithm proposed in this paper attempts to add more granularity in detection results such that, in the future, the detection results can provide information on, not only use and non-use of seat belt, but also proper/improper use cases judged by the relative position of seat belt to the detected body position.

\section{Method}

In this paper, we propose new NADS-Net architecture for simultaneous pose estimation and seat belt detection. More specifically, the main objectives are (1) to estimate 2D body posture of the driver and (if exists) the front-row passenger; and (2) to segment image pixels that correspond to seat belts. 

\subsection{Problem overview}
The driver and passenger pose estimation problem is similar to the generic 2D human pose estimation in the wild problem, in a sense that we aim to detect body key points and skeletons parsing those key points. However, there are several key differences between the driver/passenger pose estimation problem and the generic pose estimation problems as described below.

% First, we are not required to detect lower extremities given the use scenarios of IVMS systems. In terms of designing CNNs, such a difference may allow smaller receptive fields and, thus, more nimble architecture. However, in the meantime, lower extremities not visible in the video frames might also reduce the amount of contextual cues when detecting other body parts.

% \paragraph{Background/Lighting}
Most of the pose estimation models are trained and validated on publicly available data set such as MS COCO \cite{mscoco2014} and PoseTrack \cite{Andriluka2017PoseTrackAB} data sets. These data sets are, however, mostly images taken in daytime or bright indoor scenes, whereas the illumination in vehicles can vary drastically. Furthermore, in generic data sets, there is no nighttime infra-red (IR) image, hence the performance of a model trained on generic data sets is questionable in IVMS settings. This will be justified later in this paper.

On the other hand, generic data sets contain a variety of human poses whereas poses of drivers and passengers in vehicles are quite limited. Moreover, background texture and the number of people in the generic data sets are more diverse and the pixel-height of the people can also vary largely. In contrast, those quantities vary only narrowly in vehicle environments. From this observation, we hypothesize that a shallower model with lesser parameters would suffice the pose estimation task in IVMS settings. Hence, a higher computational efficiency can be achieved by reducing the neural network architecture without compromising the model performance.

% Lastly, when training deep neural network, task related data set are always required for training the model. However, considering the fact that the generic data sets contain more diverse background and variety of human poses, in-vehicle specific data set might not be necessary since training with generic data sets are already enough to cover the features of in-vehicle background and poses of driver/passenger. We will discuss this issue in the result section.

% Therefore, we hypothesize that in-vehicle human posture estimation model can be smaller in size compare to general posture estimation models since number of parameters responsible for detecting diverse backgrounds, poses, key points, and number of people are much smaller. 

\subsection{Data set}
One of the main challenges in this study was the lack of appropriate data sets. As noted above, there are many publicly available data sets for more generic human pose estimation problems, but they are not quite suited for in-vehicle monitoring purposes. Especially, we require seat belt annotations, diverse demographics, nighttime IR images, people under dynamic change of illumination as they drive, and human poses and gestures in the context of driving.

\paragraph{Data collection} We collected videos of drivers and passengers in a Volvo XC90 research vehicle through on-road driving studies. Over 7 months ranging from Spring to Winter, the total of 100 subjects consented to participate in the study in compliance to the internal review board (IRB) requirements. The subjects were randomly assigned in one of driving sessions that varied in season, weather, and time of the day. Each driving session was composed of static sessions while vehicle was at park where subjects were instructed to pose a specific set of predefined gestures, and on-road driving sessions. During the static gesture sessions, the subjects were requested to perform certain gestures and motions such as drinking, using smart phones, exercising, yawning, sneezing, leaning on the door, putting hands out the window, searching floor and the center console, adjusting sun visor, and etc. For the safety reasons, no request to perform a gesture or motion was presented to the subjects during the on-road driving sessions.

\begin{figure}
    \centering
    \begin{subfigure}[b]{\linewidth}
        \centering
        \includegraphics[width=\linewidth]{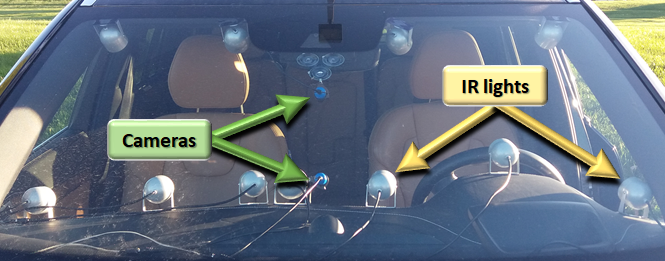}
    \end{subfigure}
    \begin{subfigure}[b]{\linewidth}
        \centering
        \includegraphics[width=\linewidth]{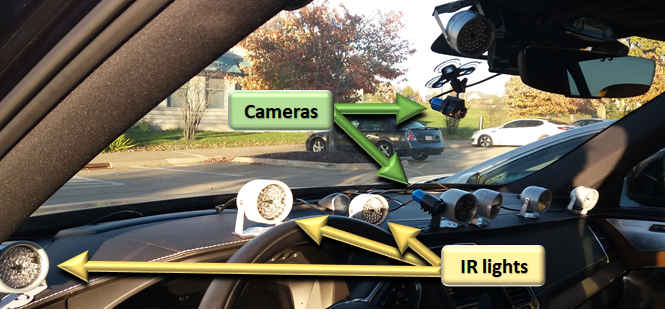}
    \end{subfigure}
    \caption{Data collection setup for this study.}
    \label{fig:volvo}
\end{figure}

For data collection, we equipped the research vehicle with IR lights and two cameras. One of the cameras was mounted below the rear view mirror and another was above the center media console. IR lights were installed on the dashboard and behind the sun visors. Figure~\ref{fig:volvo} shows how the vehicle was instrumented. 

\begin{table}[ht]
\centering
\small
    \begin{tabular}{ c | c | c | c | c | c | c }
        \toprule
            \multicolumn{2}{c|}{Sex} & \multicolumn{5}{c}{Race}\\
            \hline
            Men & Women & White & Black & Asian & Hisp. & N/A$^*$\\
            \hline
            53 & 47 & 67 & 11 & 12 & 5 & 5 \\
        \midrule
            \multicolumn{7}{c}{Age (yr.)}\\
            \hline
            10-19 & 20-29 & 30-39 & 40-49 & 50-59 & 60-69 & 70-\\
            \hline
            4 & 28 & 17 & 16 & 15 & 17 & 3\\
        \bottomrule
        \multicolumn{7}{r}{\footnotesize $^*$Not responded.}\\
    \end{tabular}
    \newline
    \vspace*{1em}
    \begin{tabular}{ c | c | c | c | c }
     \toprule
         \multicolumn{3}{c|}{Eyewear} & \multicolumn{2}{c}{Facial Hair} \\
         \hline
         None & Glasses & Sunglasses & None & Beard \\
         \hline
         47\% & 33\% & 20\% & 87\% & 13\%\\
     \midrule
         \multicolumn{3}{c|}{Clothing (Top)} & \multicolumn{2}{c}{Accessories}\\
         \hline
         Short Sleeves & Long Sleeves & Jacket/Coat & Scarf & Hat\\
         \hline
         10\% & 55\% & 35\% & 20\% & 18\%\\
     \bottomrule
    \end{tabular}
\caption{Subject statistics.}
\label{tbl:subject_statistics}
\end{table}
\paragraph{Statistics} In addition to the driving videos, we also collected demographics information of each subject such as age, sex, and race through a survey questionnaire. Additionally, researchers in this study have manually annotated videos to label clothing and accessory types. These are summarized in Table~\ref{tbl:subject_statistics}.

% Demographics statitics: total 100
% - Male = 53, Female = 47, White = 67, Black = 11, Asian = 12, Hispanic = 5, Not responded = 5.
% -	Age: 10s = 4, 20s = 28, 30s = 17, 40s = 16, 50s = 15, 60s= 17, 70s = 1, 80s = 2

% Appearance statistics: total = 214
% -	Glasses: 70, Sunglasses: 42, T-shirt: 20, Outerwear = 75, long sleeves = 118, Hat = 43, scarf = 38, Beard = 27.

It should be noted that all driving sessions were accompanied by a research staff as a safety protocol and, thus, the videos contain some repeated appearances of a few research staffs. To minimize the potential bias in the data, the researchers rotated the duty across the driving sessions. By the safety requirement, the researchers had to sit on the front passenger seat when the vehicle was in motion, but while the vehicle was at park, they moved around to different seat positions as much as possible to minimize the data bias. Moreover, researchers were asked to wear different clothing and accessories each time.

Lastly, the route of driving included a good mixture of rural roads, urban areas, and highways to diversify background and illumination.

\paragraph{Data annotation}
For each session, short video clips were selected manually and diversity in terms of subject demographics, illumination, and pose was promoted. The annotation process was done manually by human annotators. For each image, the annotators were instructed to segment all visible seat belts with a binary mask and to mark $x$ and $y$ pixel coordinates of body key points, as displayed in Figure~\ref{fig:ICCV_gt_img}, following the common convention in other publicly available pose estimation data sets such as MS COCO. We did not track the lower extremities as they were not visible in most of the frames.
%When the key points were not explicitly visible, annotators were asked to use their ``best guess'' and put an occlusion flag.
The researchers of this study conducted a final check each time the annotators submitted the job in order to assure the data quality. Annotation errors were fixed by the researchers or sent back to the annotators for rework. Sample annotation results are presented in Figure~\ref{fig:ICCV_gt_img}.
\begin{figure*}[!ht]
\centering
    \includegraphics[width=\textwidth]{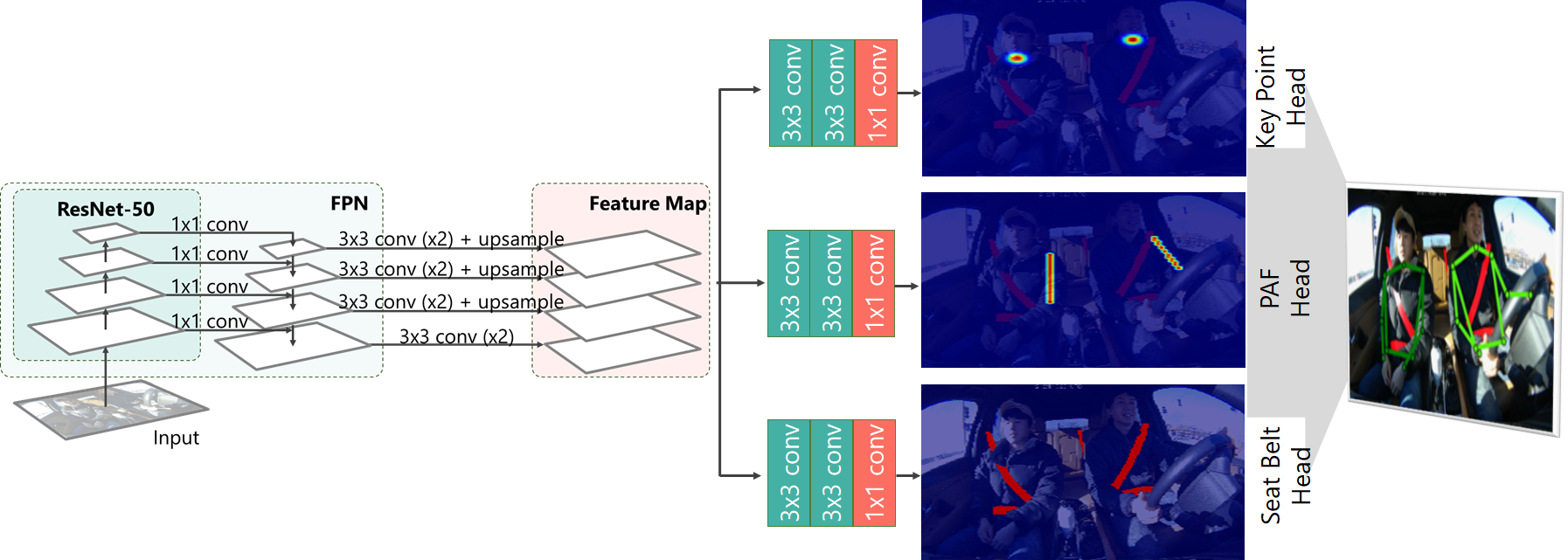}
    \caption{NADS-Net architecture.}
    \label{fig:architecture}
\end{figure*}

% \paragraph{Data Availability}
% The data set collected in this study is only for proprietary use of the sponsors {\color{red} (Is it true? Can we make it available? Just track who are using our data. Talk to Aisin.)} and not publicly available.

% \begin{figure}[!b]
% \vspace{-0.4cm}
% \centering
%     \subfigure[]{
%         \includegraphics[width=\linewidth]{figures/volvo_1.jpg}
%         \label{fig:regression}
%     }
%     \subfigure[]{
%         \includegraphics[width=\linewidth]{figures/volvo_2.jpg}
%         \label{fig:segmentation}
%     }
% \vspace{-0.4cm}
% 	\caption{\textbf{Examples of geometric data analysis.} (a) Regression of a QoI on geometry \cite{SHAPEMATTERS,gunpinar2019generative}; (b) Semantic feature segmentation \cite{Harik2017:ShapeTerra,shi2018manufacturability,he2018predicting}; (c) Point-to-point correspondence matching \cite{Sun2017a}; (d) Object detection and localization \cite{lidarimage}.}
% 	\label{fig:overview}
% \end{figure}

%\includegraphics[width=5cm]{figures/pck}\\
%\includegraphics[width=8cm]{figures/ICCV_gt_img}\\
%\includegraphics[width=5cm]{figures/Route_map}\\

% \includegraphics[width=5cm]{figures/age_histogram}
% \includegraphics[width=5cm]{figures/Height_histogram}
% \includegraphics[width=5cm]{figures/Weight_histogram}
\subsection{Model}
Our algorithm has three heads that generate key point heat maps, PAF heat maps, and a binary seat belt detection mask, sitting on top of the FPN backbone (Figure~\ref{fig:architecture}). Outputs from the first two heads are used to parse the key point instances into human skeletons. For the parsing, we employ the same PAF mechanism with the bipartite graph matching as proposed in Cao \etal \cite{cao2017paf}. However, our contribution is the architecture that generates feature maps faster and more efficient. More specifically, we reduced the six-stage architecture of \cite{cao2017paf} to a single stage architecture. Instead, we replaced the VGG \cite{Simonyan2014VeryDC} backbone with a strong multi-scale FPN backbone to speed up the inference time and to compensate the reduced staging.

The FPN backbone of NADS-Net is comprised of ResNet-50 and produces a rudimentary feature pyramid for the later detection branches. The inherent structure of ResNet can produce multi-resolution feature maps after each residual blocks, namely C2, C3, C4, and C5, which are sized $1/4$, $1/8$, $1/16$, and $1/32$ of the original input resolution, respectively. For example, for a given 384$\times$384 image input we use in the NADS-Net implementation, the ResNet-50 backbone produces four levels of feature pyramid, each sized 96$\times$96, 48$\times$48, and 24$\times$24 and 12$\times$12. Along such, the number of channels (feature maps) increases from 256 (C2) to 512 (C3), 1,024 (C4), and 2,048 (C5). These are then further convolved with 1$\times$1 convolutions, to compress the number of channels to 256. Lastly, the reduced feature pyramid further undergoes two more 3$\times$3 convolutions and an upsampling to produce a concatenated $96\times96\times512$ feature map.

Each of the detection branches employs two 3$\times$3 convolutions and a 1$\times$1 convolution to predict a pixel-wise probability distribution. For the key point head, the pixel-wise probability indicates the probability of the corresponding pixel being a certain joint. Since we are interested in detecting joints with background, the key point head produces ten such probability maps of the size $96\times96$, each of which corresponds to one of the nine joints we are interested in detecting and also background. For the PAF head, similar to \cite{cao2017paf}, we produce vector fields of size $96\times96$ which encode pairwise relationships between body joints. Lastly, the seat belt head produces a probability distribution map of a size $96\times96$ indicating the likelihood of each pixel being a seat belt. Each pixel-wise probability is then thresholded to generate a binary seat belt detection mask.

% These multi-scale feature maps provide strong representations and high resolution. Considering proven performance of the FPN in detection and segmentation tasks, this approach allows us to remove the six-stage procedure to speed up the algorithm while keeping the accuracy high. 
% Four feature maps from the Shared Feature Pyramid Backbone are passed to each of three branches, convoluted 2 times with 3x3 kernels then upsampled accordingly to fit into a concatenation step. Assumption is that the final concatenation feature map includes a multi-scale representation of intended features. After concatenation, feature map will be convoluted 2 more times and finally 1x1 convolution for the key point and paf heatmaps and segmentation.
% At the end, the model will extract three outputs simultaneously: key point head will have outputs of 10 channels and PAF head will have 16 channels. Seat belt head will have only one channel heatmap. 

\subsection{Implementation}

The proposed NADS-Net was implemented in Keras \cite{chollet2015keras} with TensorFlow \cite{tensorflow2015} backend and an NVIDIA GeForce GTX 1080 Ti GPU was used for training and testing. For the training data, 30 driving sessions out of total 50 were used. The rest were reserved for testing. When selecting 30 sessions of the training data, we manually selected half of the nighttime sessions to distribute nighttime IR images equally for both training and testing data. Rest of the training data sets were selected randomly. At the end, 10,550 images were used for training and 7,721 images were used for testing.

We pre-trained the model with MS COCO {\tt train2017} data set and the corresponding human key point annotations. Only the body key point branch and PAF branch were pre-trained. As reported in the result section, the transfer learning strategy provided a significant performance gain. Additionally, random image augmentations were applied to training images, such as rotation, scaling and vertical flipping.

For the final parsing of the skeleton, we strictly followed the implementation of Cao \etal \cite{cao2017paf}. That is, non-maximum suppression was used on the detection confidence maps which allowed the algorithm to obtain a discrete set of part candidate locations. Then, bipartite graph was used to group each person. More details are deferred to \cite{cao2017paf}.

\section{Result and discussion}
% (Comment) I have all the numbers (pckh0.5 \& pckh0.8 results for Aisin and CMU) but I don't think it's realistc to put all these data into the paper. So we need to talk about which one to put and also if it is necessary to put all the pckh for each joint. 
% {\color{red}Let's do linear regression instead of item-wise reporting. We can talk.}

% Below is the jointwise pckh evaluation
%cmu: [0.78 0.89 0.95 0.88 0.95 0.88 0.72 0.68 0.68]
%ours(MC): [0.8  0.91 0.97 0.88 0.96 0.89 0.73 0.74 0.71]
%ours(ND): [0.5  0.72 0.91 0.97 0.93 0.72 0.39 0.88 0.87]
%ours(ALL): [0.8  0.88 0.96 0.98 0.97 0.88 0.72 0.92 0.92]

\begin{figure}
\centering
    \includegraphics[width=\linewidth]{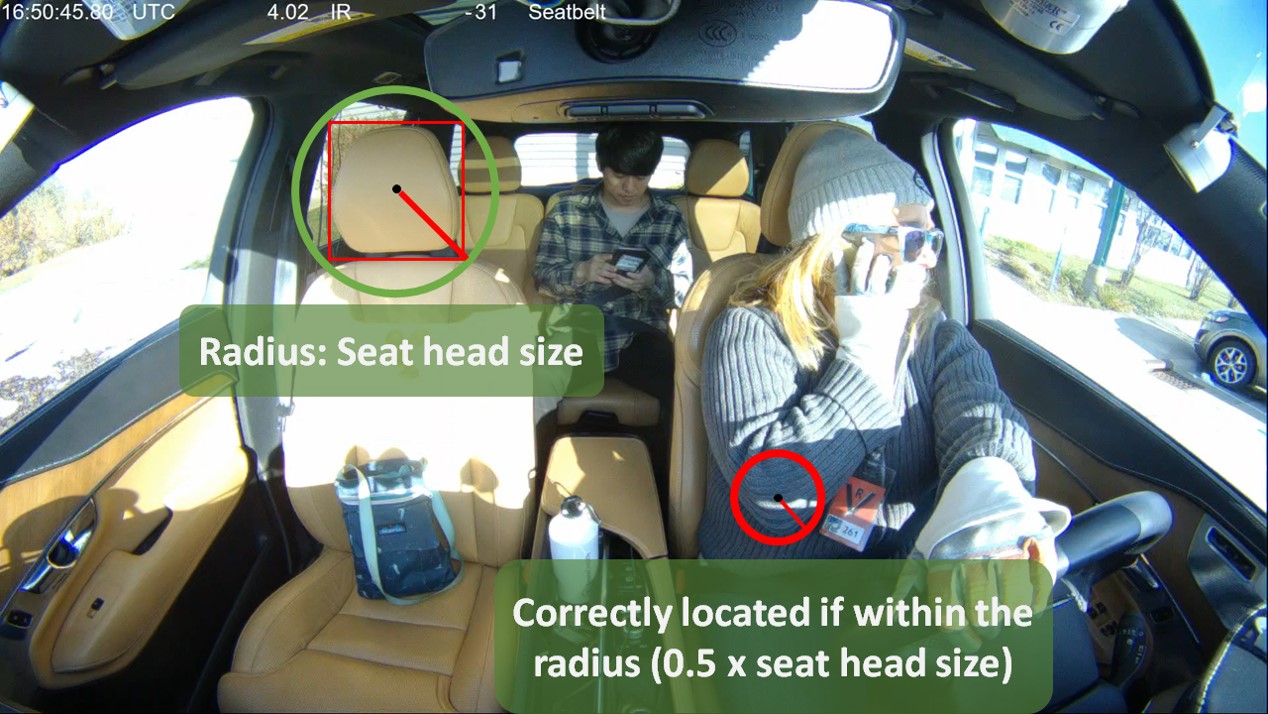}
    \caption{Modified PCKh metric (mPCKh) used for key point evaluation.}
    \label{fig:pck}
\end{figure}

We compare the NADS-Net with the PAF model \cite{cao2017paf} as a baseline. For the detection accuracy of the body key points, we employ the probability of correct key point (PCK) metric \cite{yang2012articulated} as a criterion. In typical generic human pose estimation applications, the head size of the person being estimated is used as a reference of PCK to determine the tolerance of correctness (PCKh) \cite{andriluka20142d}. This is reasonable for generic applications where the pixel heights of people vary drastically within and across images. However, for the specific in-vehicle monitoring task presented in this paper, we find such a generic way may prevent precise characterization of model performance as the head size can greatly vary depending on the spatial position of the head, while the distance from the camera to the rest of the body (for example, hands on the steering wheel) remains unchanged. To this end, we may benefit by using the headrest size as the reference of the PCK measure instead. The reason can be that, first of all, the distance from the camera to the headrest is almost the same, which allows more stable reference for PCK evaluation. Furthermore, the headrest is about the same size as the human heads, resulting the similar range of PCK values as other human pose estimation literature. This enables more intuitive interpretation of the analysis results. Therefore, we use a modified PCKh metric (mPCKh) where the diagonal length of the headrest is used as the reference (Figure~\ref{fig:pck}). It is worthwhile to note that, although the camera was fixed at the same position and angle across the entire data collection sessions, the pixel size of the headrest might change due to the seat position. However, with respect to the average diagonal length of the headrest (170 pixels), the variation is negligible (less than 10 pixels).

For the seat belt detection task, there is no baseline model available to compare. Instead, we simply report our model's sensitivity, specificity, precision, F1 score, and intersection over union (IOU). As we will discuss below, these are arguably inappropriate ways to characterize the seat belt detection accuracy, but we defer the development of a better metric to our future work.

\begin{figure}
    \centering
    \begin{subfigure}[b]{0.49\linewidth}
        \centering
        \includegraphics[width=\linewidth]{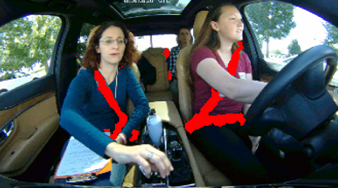}
    \end{subfigure}
    \begin{subfigure}[b]{0.49\linewidth}
        \centering
        \includegraphics[width=\linewidth]{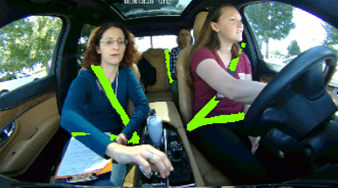}
    \end{subfigure}
    \caption{Seat belt detection result (\textit{left}) in comparison with the ground truth (\textit{right}). With human inspection, the prediction result is of good quality as it correctly marks the seat belt area. However, IOU for this particular example was 46\%, justifying the need for a better evaluation metric.}
    \label{fig:seat_eval}
\end{figure}

\subsection{Efficiency}
In terms of the total number of parameters, the baseline model requires 52,311,446 parameters while NADS-Net uses 39,334,301 parameters, which is about $25\%$ lesser despite the fact that there is an additional seat belt segmentation head. Given that the skeleton parsing method in NADS-Net is exactly the same as in \cite{cao2017paf}, the run-time difference is only a function of the model inference time. The frame rate on an Intel\textregistered\, Core$^\text{TM}$ i7-7800X 3.50 GHz CPU machine with a 32GB RAM and a NVIDIA GeForce 1080Ti GPU, the benchmark model demonstrated 12 fps while NADS-Net showed 18 fps in average. This shall not be directly interpreted as a definitive speed comparison at their maximal performance but, in fact, further optimization may dramatically change the frame rate. However, the result still serves as a weak but convincing evidence that the NADS-Net performs more efficiently than the baseline model.

\subsection{Key point detection}

We compared mPCKh scores of NADS-Net model and the baseline model for each individual key point location. As reported at the bottom of Table~\ref{tbl:pckh}, the baseline model scored the average mPCKh of 82\% over all key points while NADS-Net model scored 84\%. Unlike the baseline model, NADS-Net does not have refining stages and have fewer parameters, but PCK-wise, shows a similar or slightly better performance. This demonstrates that the multi-resolution feature pyramid produced by the FPN backbone is enough for the driver/passenger pose estimation task and can replace multiple refining stages of generic pose estimation algorithms. Some qualitative results are given in Figure~\ref{fig:ICCV_gt_img}.

\begin{table}[t]
\small
    \begin{tabular}{ c | c | c | c | c }
    \toprule
        & Cao \etal \cite{cao2017paf} & \multicolumn{3}{c}{NADS-Net (Ours)}\\
        & MC$^*$ & MC & ND$^{**}$ & All$^{***}$\\
        \hline
        Drivers       & 80\%   & 81\% &  75\%& 88\%\\
        Front Passengers &  85\% & 89\%& 78\% & 90\%\\
    \hline
        Men           & 84\% & 87\% & 79\% & 90\% \\
        Women         & 79\% & 81\% & 73\% & 88\% \\
        White         & 83\% & 86\% & 77\% & 89\% \\
        Black         & 75\% & 77\% & 70\% & 87\% \\
        Asian         & 85\% & 87\% & 80\% & 91\% \\
    \hline
    %     $\text{BMI} < 20$    &      &      &      &      \\
    %     $\text{BMI} \ge 20$  &      &      &      &      \\
    % \hline
        Glasses       & 83\% & 83\% & 75\% & 89\% \\
        Sunglasses    & 78\% & 82\% & 72\% & 85\% \\
        Short Sleeves & 84\% & 85\% & 77\% & 89\% \\
        Long Sleeves  & 85\% & 87\% & 78\% & 90\% \\
        Jacket/Coat   & 79\% & 81\% & 75\% & 88\% \\
        Scarf         & 82\% & 84\% & 78\% & 91\% \\
        Hat           & 82\% & 84\% & 77\% & 90\% \\
        Beard         & 82\% & 87\% & 81\% & 90\% \\
    \hline
        Daytime       & 85\% & 86\% & 77\% & 89\% \\
        Nighttime     & 74\% & 77\% & 75\% & 88\% \\
    \hline
        \textbf{Overall} & \textbf{82\%} & \textbf{84\%} &  \textbf{77\%} & \textbf{89\%}\\
    \bottomrule
    \multicolumn{5}{r}{\footnotesize $^{*}$ Trained with MS-COCO data set only.}\\
    \multicolumn{5}{r}{\footnotesize $^{**}$ Trained with new driving data set only.}\\
    \multicolumn{5}{r}{\footnotesize $^{***}$ Trained with both data sets combined.}
    \end{tabular}
    \caption{Accuracy evaluation with mPCKh@0.5.}
    \label{tbl:pckh}
\end{table}

\subsection{Seat belt detection}
The seat belt detection head produces a probability density function defined over the image domain, indicating the likelihood of a pixel being a seat belt. The probability distribution is then thresholded to obtain a binary seat belt segmentation mask (see Figure~\ref{fig:ICCV_gt_img}). We evaluate the quality of seat belt segmentation using five metrics: sensitivity, specificity, precision, F1 score, and IOU as reported in Table~\ref{tbl:seatbelt_eval}.

%. The test set is identical to subset that was used for key point evaluation but since ground truth seat belt mask contains very small portion of the segmentation when the seat belt is not fastened, we only evaluate the image if the corresponding ground truth mask have non-zero pixels above 0.3\% of the total number of pixels. Results are shown in Table ~\ref{tbl:seatbelt_eval}. 

The high specificity of the model indicates that the model can correctly classify non-seat belt pixels with a high confidence. However, the other metrics are poor, where the sensitivity, precision, and F1-score were 63.51\%, 63.58\%, and 63.55\%, respectively, and IOU was only 47\%. A possible interpretation of this result is that, first of all, NADS-Net model is highly conservative and predicts seat-belt only when there is a high certainty. This can be justified from the strong contrast between the sensitivity and specificity. Furthermore, even if the seat belt detection is correct, just because the predicted seat belt is thinner than the actual ground truth annotation, metrics such as sensitivity and IOU drops significantly as they penalize the thin subset of seat belt that are not detected. Lastly, we also noticed that the manual annotation of seat belt contained a few inconsistencies, which we could not resolve in this study. For example, there was an inconsistency among the annotators where some people discerned the seat belt buckles as a part of the seat belt while the others exclusively labeled the fabric part of the seat belt. These are possible sources of low sensitivity, precision, F1-score, and IOU and need to be addressed in future work.

However, more fundamentally, it is worthwhile to note the lack of suitable evaluation metrics. We inspected the seat belt segmentation results image by image and noticed that most of the error indeed comes from seat belt predicted thinner than the ground truth annotation (\eg Figure~\ref{fig:seat_eval}). A possible solution to this is to skeletonize the seat belt mask and compare the distances between the curves. Another potential solution is a metric such as optimal transport \cite{optimaltransport}. These will also be the potential venues for future study.

% For the seat belt evaluation, we lack of suitable evaluation metric. Purpose of seat belt segmentation is to realize the correct orientation of the seat belt but evaluation metrics such as sensitivity, precision, F1 scores and IOU are all pixel-wise evaluation method and hence it is not suitable evaluation method for our task. For example, even when the seat belts segmentation is well aligned with the actual seat belt, if the segmentation is too thin or thick, all 4 metrics will decrease significantly. 

% Another possible reason why the scores are low is because new data set contain inconsistent seat belt segmentation from annotators. For example, sometimes the groundtruth segmentation data set does not contain seat belt buckle and anchor segmentation. Therefore, although the model correctly segmented buckles and anchors, when groundtruth segmentation of those does not exist, all four metric scores will be low. 

\begin{table}[t]
\centering
\small
    
    \begin{tabular}{ c | c | c | c | c }
     \toprule
         Sensitivity & Specificity & Precision & F1-Score & IOU \\
         \hline
         63.51\% & 99.28\% & 63.58\% & 63.55\% & 46.57\%\\
     
     \bottomrule
    \end{tabular}
\caption{Seatbelt Evaluation}
\label{tbl:seatbelt_eval}
\end{table}

% seat belt threshold at 0.2
% gt image threshold is 0.3%
%total # of testing imgs: 7721
% total # imgs calculated: 6108
%sensitivity = 0.64
%specificity = 0.99
%precision = 0.64
%F1 =  0.64
%iou =  0.47

% seat belt threshold at 0.2
% when threshold is 1%
%total # of testing imgs: 7721
% total # imgs calculated: 3189
%sensitivity = 0.66
%specificity = 0.99
%precision = 0.68
%F1 =  0.67
%iou =  0.50

%\begin{figure*}
%\centering
%    \includegraphics[width=\linewidth]{figures/result_sample.png}
    
%    \caption{sample results. \color{red} ADD EXPLANATIONS. DRIVERS and Passengers need to be the consistent color. Sometimes, the front row passenger is green, sometimes the driver is green. This needs to be fixed. Also, it's a bit confusing because we said we are interested in detecting only the driver and passenger. However, the figure demonstrates the back passengers as well.}
%    \label{fig:ICCV_result_sample}
%\end{figure*}

\subsection{Appearances}
% All evaluated images contain person key point annotations and various labels such as gender, race, and clothing. Gender and race information are collected from surveys that participants filled out but clothing and accessory labels are classified manually from a researcher by reviewing each images in the new data set. 

\paragraph{Performance on different demographics}
In Table~\ref{tbl:pckh}, evaluation of the model performance on different demographic parameters is reported. For women, the overall performance was lower than men for all four experiments---the baseline model trained only on MS COCO data set; NADS-Net model trained only on MS COCO data set; NADS-Net model trained only on the new driving data set; and NADS-Net model pretrained on MS COCO data set and then transferred to the new driving data set. Considering the fact that women-to-men ratio was 1:1, one hypothesis we can derive from this observation is that the appearance variance among women is larger than men, because of larger diversities in hairstyle, accessories, clothing patterns, etc. among female populations. Therefore, it is more advisable to include more female subjects in data collection in the future.

Race-wise, the model performance was slightly better on Asian populations followed by white populations. Performance on people with darker skin tone was noticeably lower, reconfirming the bias of computer vision data sets and algorithms as pointed out by Zou and Schiebinger \cite{AIracist}. We believe the new driving data set collected in this study also can suffer from the same bias. The primary reason was the geographical location where the study was conducted, whose population was predominantly white. Furthermore, coincidentally 60\% of black subjects participated in the study during the nighttime, which could also worsen the performance evaluation on this demographic group. The future work, therefore, needs to include more subjects with darker skin tones, to overcome the bias in performance and a more rigorous and controlled analysis.

% Performance on White is same or slightly better than reported average score and performance on Asian group was 1\% or 3\% higher than White group. Performance difference between White and Asian is small while noticeable difference is shown from the Black labeled group. Reported scores on Black population are 7\% lower than the overall score for all evaluated model except for NADS-Net(All). The performance of NADS-Net(All) on Black population was only 2\% lower than the overall score. Reported score difference may indicate the data sets are lack of Black populations and thus generates performance biases. 

\begin{figure}
    \centering
    \begin{subfigure}[b]{0.49\linewidth}
        \centering
        \includegraphics[width=\linewidth]{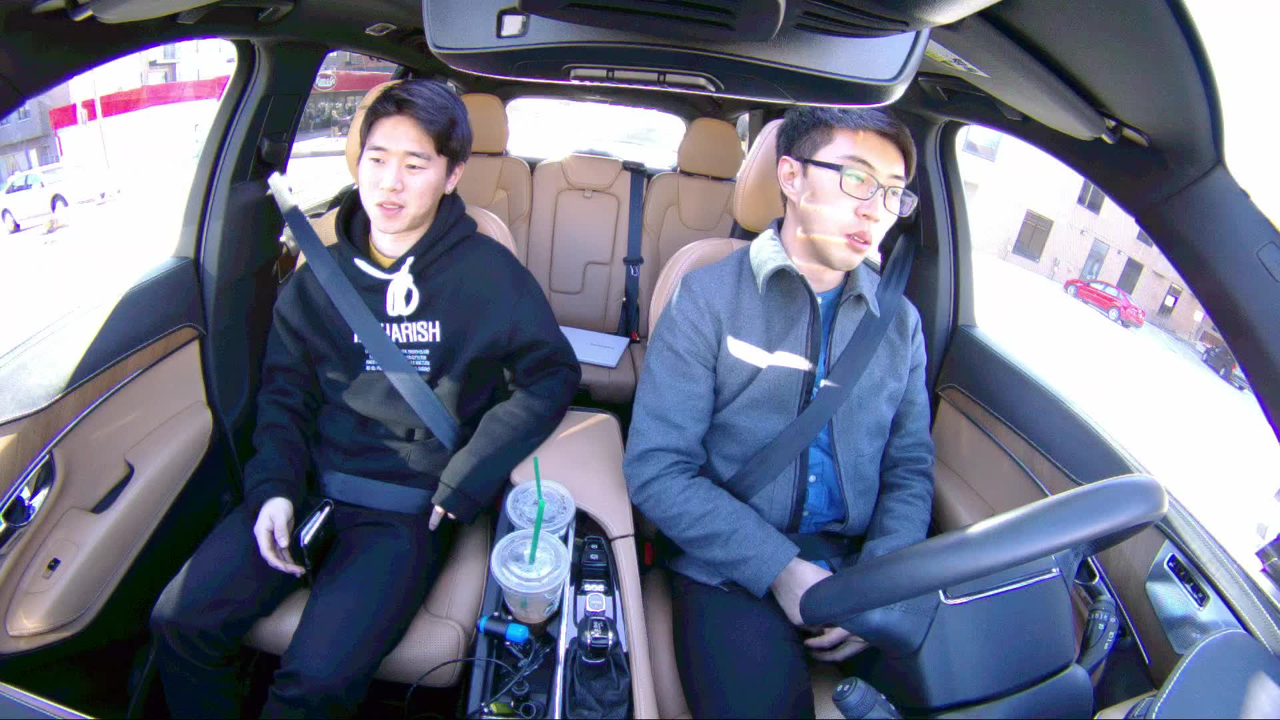}
    \end{subfigure}
    \begin{subfigure}[b]{0.49\linewidth}
        \centering
        \includegraphics[width=\linewidth]{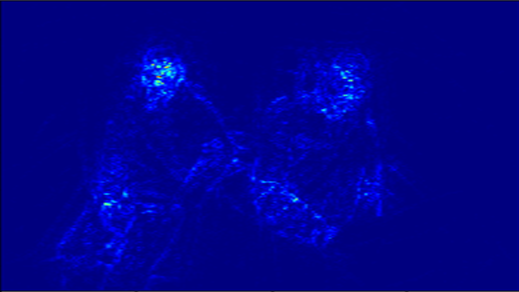}
    \end{subfigure}
    \caption{Saliency map visualization \cite{simonyan2013deep} for the `right wrist' class. Note visual cues from the face have significant contribution to the prediction.}
    \label{fig:saliency_map}
\end{figure}
\paragraph{Clothing/accessories}
Table~\ref{tbl:pckh} reports the model performance on nine different clothing/accessory categories. The numbers reveal that there is no significant influence of clothing/accessories, except for sunglasses. We failed to reach a convincing explanation on this, but a weak hypothesis on this might be that many of the visual cues to a CNN-based pose estimation model come from the facial area. As demonstrated in the saliency map visualization in Figure~\ref{fig:saliency_map}, even for the detection of right wrist, for example, which is relatively far from the face, we can notice the detection relies largely on the visual cues coming from the face, not just the wrist and the arm area. Although this is beyond the scope of this paper, it should be worth investigating this in future work, to deepen our understanding of how pose estimation models perceive and process the visual cues.

Another noticeable fact was that the model performance was poor on people wearing jackets/coats, when the model was trained on MS COCO data set only, but the performance improved significantly when transferred to the new driving data set. This might suggest that MS COCO data set is biased to lighter clothing but confirming this hypothesis by examining the data set image by image should be beyond the objective of this study.

\paragraph{Illumination}
We could observe a significant drop of performance in nighttime data when the model was trained on MS COCO data set only. This could be easily improved by transferring the model to the new driving data set. We can conclude that there is not much of fundamental differences between daytime images and nighttime IR images in terms of visual cues available for the detection of body parts. Instead, the drop of performance mostly comes from the lack of nighttime data in MS COCO data set, not an inherent deficiency of CNNs.

% With daytime images only, Cao\cite{cao2017paf} scored 85\% mPCKh and NADS-Net(MC) scored 86\%. However, the mPCKh score drops about 10\% for both models when evaluated on nighttime images only. When NADS-Net is trained with the new data set only or fine-tuned after training with MS COCO data set, mPCKh score difference between daytime and nighttime is only 1\% to 2\%. Accuracy drop is natural for models trained with MS COCO since MS COCO data set does not contain nighttime images. This suggest the necessity of collecting new data set including nighttime images for in-vehicle posture estimation task. 

% \subsection{PAF - neck thing}

\section{Conclusion and future work}

In this paper, we proposed a new CNN architecture called NADS-Net for the purpose of driver and passenger pose estimation and seat belt detection in vehicles. NADS-Net is capable of estimating body pose together with seat belt segmentation with the similar accuracy than the state of the art baseline \cite{cao2017paf}, while requiring fewer parameters and demonstrating a faster inference time. We broke the performance down and provided in-depth analyses in different aspects, including sex, race, clothing, and illumination. These results may provide practical insights to future academic research and to industrial product development.

For the future work, one of the clear challenges is the bias of data set. One trivial solution could be to enlarge the scale of data collection study by including more diverse group of subjects and other imaging parameters. However, practically, this may not be viable in many aspects. To this end, one potential direction we are exploring towards is the creation of a synthetic data set and randomizing imaging conditions virtually. In addition, the current status of our work is limited to a frame-by-frame detection, while, arguably, it is more desirable to take temporal aspects (\eg optical flow) into account as in recent works such as \cite{videopose3d}.
%Finally, higher level information such as activity recognition, emotion/state recognition, and movement forecasting could also be generated, building upon the work introduced in this paper.

% According to our evaluation, although NADS-Net trained only with MS COCO \cite{mscoco2014} data set achieves high accuracy, fine-tuning the model increases 5\% more accuracy which reveals that domain specific data set is needed to further improve the performance. Accuracy increased relatively high for nighttime images which indicates the necessity of collecting data set with nighttime scenes.

% However, collecting data set is always a bottle neck for developing in-vehicle monitoring system. To overcome this issue without collecting another data set, we plan to investigate feasibility of augmentation techniques such as applying domain randomization, random contrast and intensity, and Gaussian blur in the future to make the model focus more on the shape of human body and seat belt instead of color or intensity of the image. 

% Moreover, we plan to investigate other factors such as combination of two or more labels that may contribute to detection failures. Also, we will develop a deep classifier that can classify driver activity recognition from predicted poses and identify whether the driver is properly buckled or not from seat belt segmentation result. 

{\small
\bibliographystyle{ieee}
\bibliography{main}
}

\end{document}